\documentclass[12pt]{article}

\usepackage{amsthm}
\usepackage{mathtools}
\usepackage[colorlinks,citecolor=blue,urlcolor=blue,filecolor=blue,backref=page]{hyperref}
\usepackage[nottoc]{tocbibind}
\usepackage{float}

\usepackage[utf8]{inputenc} 
\usepackage[T1]{fontenc}    
\usepackage{booktabs}       
\usepackage{amsfonts}       
\usepackage{nicefrac}       
\usepackage{microtype}      
\usepackage{listings}
\usepackage{subfigure}
\usepackage{epsfig}

\usepackage{amsmath}
\usepackage{graphicx,psfrag,epsf}
\usepackage{enumerate}
\usepackage{natbib}
\usepackage{algorithm}
\usepackage{algorithmic}
\usepackage{amssymb}
\usepackage{setspace}
\usepackage{fancyhdr}

\theoremstyle{plain}
\newtheorem{theorem}{Theorem}[section]

\theoremstyle{definition}
\newtheorem{definition}[theorem]{Definition}

\newcommand{\vect}[1]{\mathbf{#1}}

\usepackage{url} 

\newcommand{\blind}{1}

\renewcommand{\baselinestretch}{1.5}

\begin{document}

\def\spacingset#1{\renewcommand{\baselinestretch}%
{#1}\small\normalsize} \spacingset{1}

\doublespacing


\if1\blind
{
  \title{A case study of Empirical Bayes in User-Movie Recommendation system}
  \author{Arabin Kumar Dey, Raghav Somani and Sreangsu Acharyya\\
    Department of Mathematics, IIT Guwahati\\   
    Microsoft Research India\\ Microsoft Research India
  } 

\maketitle

} \fi

\if0\blind
{
  \bigskip
  \bigskip
  \bigskip
  \begin{center}
    {\LARGE\bf A case study of Empirical Bayes in User-Movie Recommendation system}
\end{center}
  \medskip
} \fi

\bigskip
\begin{abstract}

 In this article we provide a formulation of empirical bayes described by \cite{atchade2011computational} to tune the hyperparameters of priors used in bayesian set up of collaborative filter. We implement the same in MovieLens small dataset.  We see that it can be used to get a good initial choice for the parameters.  It can also be used to guess an initial choice for hyper-parameters in grid search procedure even for the datasets where MCMC oscillates around the true value or takes long time to converge.        

\end{abstract}

\noindent%
\vfill

\newpage
\spacingset{1.45} 

\lhead{}
\rhead{EB - Recommendation system}

\section{Introduction}
\label{sec:intro}

  Some of the most successful realizations of latent factor models are based on Matrix Factorization. In its basic form, Matrix Factorization characterizes both items and users by real vectors of factors inferred from item rating patterns represented by $k$ dimensional vectors $\vect{u}_i$ and $\vect{v}_j$ corresponding to $i^{th}$ user and $j^{th}$ item respectively. High correspondence between item and user factors leads to a recommendation. The most convenient data is high-quality explicit feedback, which includes explicit input by users regarding their interest in the items. We refer the explicit user feedback as ratings. Usually, explicit feedback comprises a sparse matrix $\vect{M}$, since any single user is likely to have rated only a very small percentage of possible items.
Characterizing the feedback linearly, it approximates the ratings $m_{ij}$ as the dot product of $\vect{u}_i$ and $\vect{v}_j$ such that the estimate $\hat{m}_{ij} = \vect{u}_i^T.\vect{v}_j$. The major challenge is computing the mapping of each item and user to factor vectors $\vect{u}_i$, $\vect{v}_j \in \mathbb{R}^k$.  This approach of approximating the ratings $m_{ij}$ is known as collaborative filter (CF). 

 Collaborative Filter (\cite{su2009survey}, \cite{liu2009probabilistic}) required to have the ability to deal with highly sparse data, to scale with the increasing number of users of items, to make satisfactory recommendations in a short time-period, and to deal with other problems like synonymy (the tendency of the same or similar items to have different names), shilling attacks, data noise and privacy protection problems.  Usual CF does not bother to maintain the order of the ratings, which leads to stability.  Also getting a good set of hyperparameters is also important and computationally very expensive.  In this paper we mainly deal with the last problem that we have mentioned i.e. getting a good set of hyperparameters.      

  Bayesian model is used in many papers related to collaborative filter (\cite{chien1999bayesian}, \cite{jin2004bayesian}, \cite{xiong2010temporal}, \cite{yu2002collaborative}).  \cite{ansari2000internet} propose a bayesian preference model that statistically involves several types of information useful for making recommendations, such as user preferences, user and item features and expert evaluations.  They use Markov Chain Monte Carlo method for sampling based inference, which involve parameter estimation from full conditional distribution of parameters.  They achieved better performance than pure collaborative filtering.  \cite{salakhutdinov2008bayesian} discussed fully bayesian treatment of the Probabilistic Matrix Factorization (PMF) model in which model capacity is controlled automatically by integrating over all model parameters and hyperparameters.  They have tuned even hyperparameters within the Gibbs sampler framework.  However no paper uses empirical bayes procedure described by \cite{atchade2011computational} in this context.  In this paper we make a case study of this approach in Probabilistic Matrix Factorization context using different Movie Lens dataset.

 We organize the paper as follows.  In section 1 we provide the overview of naive Matrix Factorization algorithm and show some experimental results on it. In section 2 we formulate hyper parameter selection through empirical bayes method in this context. Some numerical experiment with conventional data set is discussed in section 3.  We conclude the paper in section 4.  

\section{Learning latent features}

\begin{definition}\label{abc1}
$n$ := Number of unique users,\\
$p$ := Number of unique items,\\
$k$ := Number of latent feature,\\
$\vect{M}$ := Sparse rating matrix of dimension $(n \times p)$ where the $(i, j)^{th}$ element of the rating $m_{ij}$ is given by user $i$ to item $j$.\\
$\vect{U}$ := The user feature matrix of dimension $(n \times k)$ where row $i$ represents the user feature vector $\vect{u}_i$.\\
$\vect{V}$ := The item feature matrix of dimension $(p \times k)$ where row $j$ represents the item feature vector $\vect{v}_j$.\\
$\mathcal{L}$ := The loss function which is to be minimized.\\
$\lambda_1$ \& $\lambda_2$ := User and item hyper-parameters in the regularized Loss function $\mathcal{L}$.\\
$|| \cdot ||_F$ := Frobenius norm for matrices.\\
$\kappa$ := The set of user-item indices in the sparse rating matrix $\vect{M}$ for which the ratings are known.\\
$\kappa_i$ := The set of indices of item indices for user $i$ for which the ratings are known.\\
$\kappa_j$ := The set of indices of user indices for item $j$ who rated it.
\end{definition}

  To learn the latent feature vectors $\vect{u}_i$ and $\vect{v}_j$, the system  minimizes the regularized loss on the set of known ratings.

\begin{equation}\label{eqn1}
\mathcal{L}(\vect{M}; \vect{U},\vect{V},\lambda_1, \lambda_2) = \frac{1}{|\kappa|}\sum_{i,j \in \kappa} (m_{ij} - \vect{u}_i^T.\vect{v}_j)^2 + \lambda_1||\vect{U}||_F^2 + \lambda_2||\vect{V}||_F^2
\end{equation}

  This Loss function is a biconvex function in $\vect{U}$ and $\vect{V}$ and can be iteratively optimized by regularized least square methods keeping the hyper-parameters fixed.

  The user and item feature matrices are first heuristically initialized using normal random matrices with iid. entries such that the product $\vect{U}.\vect{V}^T$ has a mean of 3 and variance 1.\\
The iteration is broken into 2 steps until test loss convergence.
The first step computes the regularized least squares estimate for each of the user feature vectors $\vect{u}_i$ from their known ratings.\\
The second step computes the regularized least squares estimate for each of the item feature vectors $\vect{v}_j$ from their known ratings.\\

The first step minimizes the below expression keeping $\vect{V}$ fixed.
\begin{equation} \label{eqn2}
|| \vect{M}_{i,\kappa_i} - \vect{V}_{\kappa_i}.\vect{u}_i||^2 + \lambda_1||\vect{u}_i||^2 \qquad \forall i = 1, 2, \ldots, n.
\end{equation}

The second step minimizes the below expression keeping $\vect{U}$ fixed.
\begin{equation}\label{eqn3}
|| \vect{M}_{\kappa_j,j} - \vect{U}_{\kappa_j}.\vect{v}_j||^2 + \lambda_2||\vect{v}_j||^2 \qquad \forall j = 1, 2, \ldots, p.
\end{equation}

The normal equations corresponding to the regularized least squares solution for user feature vectors are
\begin{equation}\label{eqn4}
(\vect{V}_{\kappa_i}^T.\vect{V}_{\kappa_i} + \lambda_1 \vect{I}_{k}).\vect{u}_i = \vect{V}_{\kappa_i}^T.\vect{M}_{i,\kappa_i} \qquad i = 1, 2, \ldots, n
\end{equation}
The normal equations corresponding to the regularized least squares solution for item feature vectors are
\begin{equation}\label{eqn4}
(\vect{U}_{\kappa_j}^T.\vect{U}_{\kappa_j} + \lambda_2 \vect{I}_{k}).\vect{v}_1 = \vect{U}_{\kappa_j}^T.\vect{M}_{\kappa_j,j} \qquad i = 1, 2, \ldots, n
\end{equation}

  Iteratively minimizing the loss function gives us a local minima (possibly global).

\subsection{Well-known Experiments}

\begin{figure}[!h]
	\includegraphics[width=0.48\textwidth]{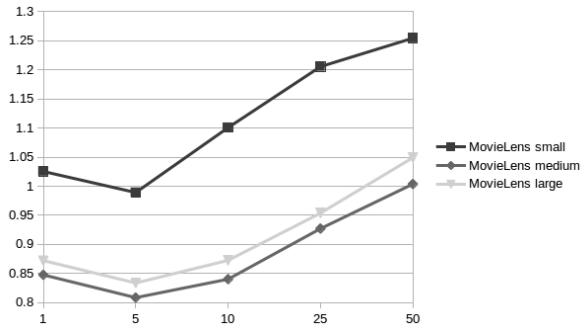}
	\caption{k vs RMSE of Matrix Factorization algorithm on Test data}
	\label{fig:1}
\end{figure}

  The benchmark datasets used in the experiments were the 3 Movie Lens dataset. They can be found at \url{http://grouplens.org/datasets/movielens/}.  The experimental results are shown in Table-\ref{table:1} below.  We see from Table-\ref{table:1} that RMSE is decreasing as sample size increases.  Figure-\ref{fig:1} shows the plot of number of latent factor ($k$) versus RMSE of matrix factorization.  Minimum RMSE is attained at $k = 5$ for all the datasets.
    
\begin{table}[H]
\begin{center}
	\begin{tabular}{|c|c|c|c|c|}
		\hline
		Dataset & n & p & Ratings & Minimum RMSE \\ \hline
		MovieLens small & 751 & 1616 & 100,000 & 0.989 \\
		MovieLens medium & 5301 & 3682 & 1,000,000 & 0.809 \\
		Movielens large & 62007 & 10586 & 10,000,000 & 0.834 \\
		\hline
	\end{tabular}	
\end{center}
\caption{RMSE of Matrix Factorization algorithm on Test data}
\label{table:1}
\end{table}

 The naive Matrix Factorization algorithm discussed in this section has certain pros and cons.  It is possible to represent each user and item into a feature vectors that can further be used into several individualistic analysis.  It tries to approximate the ratings in an iterative least square manner. Since the loss function is non-convex but still it is bi-convex, we have an iterative algorithm that results in a good local minima.  In the negative aspect of the algorithm, value fitting approach does not bother about maintaining the order of the ratings, which leads to instability.  The algorithm is sensitive with respect to hyper-parameters. So getting a good set of hyper-parameters is important and computationally very intensive.

\section{Hyper parameter selection through empirical bayes method}

 In this section we provide a brief sketch of the algorithm developed by \cite{atchade2011computational} for choosing hyper parameter through Empirical Bayes method.   Suppose that we observe a data $y \in Y$ generated from $f_{\theta, \lambda}(y)$. $f_{\theta, \lambda}(y)$ is the conditional distribution of $y$ given that the parameter takes the value $(\theta, \lambda) \in \Theta \times \Lambda$. $\lambda$ is treated as a hyper-parameter and assume that the conditional distribution of the parameter $\theta$ given $\lambda \in \Lambda$ is $\pi(\theta | \lambda)$. Therefore the joint distribution of $y, \theta$ given $\lambda$ is thus
\begin{equation}
\pi(y, \theta \mid \lambda) = f_{\theta, \lambda}(y)\pi(\theta \mid \lambda)
\end{equation}
The posterior distribution of $\theta$ given $y, \lambda$ is then given by
\begin{equation}
\pi(\theta \mid y, \lambda) = \frac{\pi(y, \theta \mid \lambda)}{\pi(y \mid \lambda)}
\end{equation}
where $\pi(y \mid \lambda)$ = $\int \pi(y, \theta \mid \lambda) d\theta$.\\

 The idea is to estimate $\lambda$ using the data $y$.  This estimate is typically taken as the maximum likelihood estimate $\hat{\lambda}$ of $\lambda$ given $\theta$.
\begin{equation}
  \hat{\lambda} = \arg\max\ \ \pi(y \mid \lambda)
\end{equation}

 The marginal distribution $\pi(y \mid \lambda) = \int \pi(y, \theta \mid \lambda)d\theta$ is not available in closed form making the maximum likelihood estimation computationally challenging.  In these challenging cases, empirical Bayes procedures can be implemented using the EM algorithm as proposed by \cite{casella2001empirical}.  This leads to a two-stage algorithm where in the first stage, an EM algorithm is used (each step of which typically is requiring a fully converged MCMC sampling from $\pi(\theta \mid y,\lambda)$) to find $\hat{\lambda}$ and in a second stage, a MCMC sampler is run to sample from $\pi(\theta \mid y, \hat{\lambda})$.  We use $\nabla_x$ to denote partial derivatives with respect to $x$.

  Let us define $l(\lambda \mid y) = \log \pi(y \mid \lambda)$ as the marginal log-likelihood of $\lambda$ given $y$ and define $h(\lambda \mid y) = \nabla_{\lambda} l(\lambda \mid y)$ its gradient. From here we have	
\begin{eqnarray} h(\lambda \mid y) & = & \nabla_{\lambda} l(\lambda \mid y) \nonumber\\
	& = & \int \frac{\partial}{\partial \lambda}\log \lbrack f_{\theta, \lambda}(y) \pi(\theta \mid \lambda) \rbrack \pi(\theta \mid y,\lambda)d\theta \nonumber\\
	& = & \int H(\lambda, \theta) \pi(\theta \mid y,\lambda)d\theta
\end{eqnarray}	
where	
\begin{equation}
	H(\lambda, \theta) := \nabla_{\lambda} \log (f_{\theta, \lambda}(y)\pi(\theta \mid \lambda)) \label{H}
\end{equation}

 In many cases, the likelihood does not depend on the hyper-parameters so that the function $H$ simplifies further to	
\begin{equation}
	H(\lambda, \theta) = \nabla_{\lambda} \log \pi(\theta \mid \lambda)
\end{equation}

 Therefore the paper proposes to search for $\displaystyle \hat{\lambda} = \arg\max_{\lambda} \pi(y \mid \lambda)$ by solving the equation $h(\lambda \mid y) = 0$. If $h$ is tractable then this equation can be easily solved analytically using iterative methods.  For example the gradient method would yield an iterative algorithm of the form
\begin{equation}
	\lambda^{'} = \lambda + ah(\lambda \mid y)
\end{equation}
for a step size $a > 0$. If $h$ is intractable, we naturally turn to \textbf{stochastic approximation algorithms}.

  Suppose that we have at our disposal for each $\lambda \in \Lambda$, a transition kernel $P_{\lambda}$ on $\Theta$ with invariant distribution $\pi(\theta \mid y, \lambda)$. We let $\{ a_n, n\geq 0 \}$ be a non-increasing sequence of positive numbers such that	
\begin{equation}
	\lim_{n \to \infty} a_n = 0 \qquad \sum a_n < \infty \qquad \sum a_n^2 < \infty
\end{equation}

 Finally, The Stochastic approximation algorithm proposed to maximize the function $l(\lambda \mid y)$ is as follows.
	\begin{itemize}
		\item Generate $\theta_{n+1} = P_{\lambda_n}(\theta, \cdot )$.
		\item Calculate $\lambda_{n+1} = \lambda_n + a_n H(\lambda_n, \theta_{n+1})$
	\end{itemize}

  Here we approximate $h(\lambda \mid y)$ by $H(\lambda, \theta)$.  This is iteratively done until the convergence of hyper-parameters.  The transition kernel $P_{\lambda}$ can be the MCMC sampler that samples from the distribution $\pi(\theta \mid y,\lambda)$. And the sequence $\{ a_n, n\geq 0 \}$ that works reasonably well is the $\frac{a}{n}$ for some $a > 0$.
	
\section{Empirical Bayes on User-Movie set up}

 Given the loss function in equation (\ref{eqn1}), we get the corresponding Boltzmann distribution,  
\begin{equation}\label{eqn_bayes}
\pi(\vect{U}, \vect{V} \mid M, \lambda_1, \lambda_2) \propto \exp \{ -\frac{1}{|\kappa|}\sum_{i,j \in \kappa} (m_{ij} - \vect{u}_i^T.\vect{v}_j)^2 \} \exp \{ -\lambda_1||\vect{U}||_F^2 \} \exp \{ -\lambda_2||\vect{V}||_F^2 \}
\end{equation}

 This energy based approach also provide us a nice bayesian connection which we exploit in selection of hyper-parameters.  Clearly equation-\ref{eqn_bayes} provides the posterior of $\vect{U}$, $\vect{V}$ where prior variances $\frac{1}{\lambda_1}$ and $\frac{1}{\lambda_2}$ connects the hyper-parameters in regularized loss function $\mathcal{L}$.  Clearly parameter set $\theta = (\vect{U}, \vect{V}).$  

  From the above the posterior, it is to easy to figure out that likelihood function or conditional distribution of observed data $\vect{M}$ given $\vect{U}$, $\vect{V}$, $\lambda_1$ and $\lambda_2$ is
\begin{equation}
f_{\theta}(m_{ij}) \propto \exp \{ -\frac{1}{|\kappa|}\sum_{i,j \in \kappa} (m_{ij} - \vect{u}_i^T.\vect{v}_j)^2 \}
\end{equation}
and the priors on $\vect{U}$ and $\vect{V}$ given $\lambda_1$ and $\lambda_2$ are
\begin{eqnarray}
\pi(\vect{U}, \vect{V} \mid \lambda) &=& \pi(\vect{U} \mid \lambda_1) \pi(\vect{V} \mid \lambda_2) \nonumber\\
&=& \exp \{ -\lambda_1||\vect{U}||_F^2 \} \exp \{ -\lambda_2||\vect{V}||_F^2 \}
\end{eqnarray}

  We obtain the random samples of $\vect{U}$ and $\vect{V}$ in every iteration from the distribution $\pi(\vect{U}, \vect{V} \mid \vect{M},\lambda_1, \lambda_2)$ using the Metropolis-Hastings algorithm.

\subsubsection{Stochastic approximation algorithm in User-Movie set up}

 First step of stochastic approximation algorithm is to draw sample from transition kernel $P_{\lambda}$ which is a MCMC sampler that samples from the distribution $\pi(\theta \mid y, \lambda)$.  We use Metropolis-Hastings algorithm to obtain a sequence of random samples from  $\pi(\vect{U}, \vect{V} \mid \vect{M}, \lambda_1, \lambda_2)$.  Let $\vect{U}^{(i)}$ and $\vect{V}^{(i)}$ be the current iterates of the iteration sequence and $q(\vect{U}, \vect{V} \mid \vect{U}^{(i)}, \vect{V}^{(i)})$ be the proposal distribution.  The algorithmic steps to get the next iterate is
\begin{itemize}
	\item Sample $(\vect{U}^*, \vect{V}^*) \sim q(\vect{U}, \vect{V} \mid \vect{U}^{(i)}, \vect{V}^{(i)})$.
	\item Calculate the acceptance probability
	\begin{equation}
	\rho((\vect{U}^{(i)}, \vect{V}^{(i)}),(\vect{U}, \vect{V})) = \min \bigg\{ 1 , \frac{\pi(\vect{U}^*,\vect{V}^* \mid \vect{M},\lambda_1^{(i)}, \lambda_2^{(i)}) q(\vect{U}^{(i)}, \vect{V}^{(i)} \mid \vect{U}^*, \vect{V}^*)}{\pi(\vect{U}^{(i)}, \vect{V}^{(i)} \mid \vect{M},\lambda_1^{(i)}, \lambda_2^{(i)})q(\vect{U}^*, \vect{V}^* \mid \vect{U}^{(i)}, \vect{V}^{(i)})} \bigg\}
	\end{equation}
	
 We assume $q$ as a symmetric proposal distribution, the acceptance probability reduces to	
\begin{equation}
	\rho((\vect{U}^{(i)}, \vect{V}^{(i)}), (\vect{U}, \vect{V})) = \min \bigg\{ 1 , \frac{\pi(\vect{U}^*, \vect{V}^* \mid \vect{M},\lambda_1^{(i)}, \lambda_2^{(i)}) }{\pi(\vect{U}^{(i)}, \vect{V}^{(i)} \mid \vect{M}, \lambda_1^{(i)}, \lambda_2^{(i)})} \bigg\} \label{acc_prob}
\end{equation}	
 Using Equation-\ref{acc_prob}, the expression for acceptance probability neatly reduces to	
\begin{equation}
	\rho((\vect{U}^{(i)}, \vect{V}^{(i)}),(\vect{U}, \vect{V})) = \min \bigg\{ 1 , \frac{\exp \{ -\mathcal{L}(\vect{U}^*, \vect{V}^*, \vect{M}, \lambda_1^{(i)}, \lambda_2^{(i)}) \}}{\exp \{ -\mathcal{L}(\vect{U}^{(i)}, \vect{V}^{(i)}, \vect{M}, \lambda_1^{(i)}, \lambda_2^{(i)}) \}} \bigg\}
\end{equation}
	and further to
\begin{equation}
	\rho((\vect{U}^{(i)}, \vect{V}^{(i)}), (\vect{U}, \vect{V})) = 	\min \bigg\{ 1 , e^{ -\mathcal{L}(\vect{U}^*, \vect{V}^*, \vect{M}, \lambda_1^{(i)}, \lambda_2^{(i)}) + \mathcal{L}(\vect{U}^{(i)}, \vect{V}^{(i)}, \vect{M}, \lambda_1^{(i)}, \lambda_2^{(i)}) } \bigg\}
\end{equation}
	\item Set $\vect{U}^{(i+1)} = \vect{U}^*$ and $\vect{V}^{(i+1)} = \vect{V}^*$ with probability $\rho((\vect{U}^{(i)}, \vect{V}^{(i)}),(\vect{U}, \vect{V}))$, otherwise $\vect{U}^{(i+1)} = \vect{U}^{(i)}$ and $\vect{V}^{(i+1)} = \vect{V}^{(i)}$.
\end{itemize}

 According to the algorithm, we sample one set of $\vect{U}$ and $\vect{V}$ using Metropolis-Hasting algorithm and use them to update hyper-parameters in the next step.

\begin{equation}
\lambda_1^{(i+1)} = \lambda_1^{(i)} + a_n H(\lambda_1^{(i)},(\vect{U}^{(i+1)},\vect{V}^{(i+1)}))
\end{equation}
and
\begin{equation}
\lambda_2^{(i+1)} = \lambda_2^{(i)} + a_n H(\lambda_2^{(i)},(\vect{U}^{(i+1)},\vect{V}^{(i+1)}))
\end{equation}
From the definition of $H$ in Equation-\ref{H}, we have
\begin{eqnarray}
H(\lambda_1^{(i)},(\vect{U}^{(i+1)},\vect{V}^{(i+1)})) &=& \nabla_{\lambda_1} \log \pi(\vect{U}^{(i+1)} \mid \lambda_1) \nonumber\\
&=& \nabla_{\lambda_1} \{ -\lambda_1 ||\vect{U}^{(i+1)}||_F^2 \} \nonumber\\
&=& -||\vect{U}^{(i+1)}||_F^2
\end{eqnarray}
similarly
\begin{eqnarray}
H(\lambda_2^{(i)},(\vect{U}^{(i+1)},\vect{V}^{(i+1)})) &=& \nabla_{\lambda_2} \log \pi(\vect{V}^{(i+1)} \mid \lambda_2) \nonumber\\
&=& \nabla_{\lambda_2} \{ -\lambda_2 ||\vect{V}^{(i+1)}||_F^2 \} \nonumber\\
&=& -||\vect{V}^{(i+1)}||_F^2
\end{eqnarray}

 In this case the sequence $\{a_n,  n > 0 \}$ is chosen as $\frac{a}{n}$ for a suitable choice of $a$.  Therefore, the hyper-parameter updates can be given by 
\begin{equation}
\lambda_1^{(i+1)} = \lambda_1^{(i)} - \frac{a}{n}||\vect{U}^{(i+1)}||_F^2 \label{lam1}
\end{equation}
and
\begin{equation}
\lambda_2^{(i+1)} = \lambda_2^{(i)} - \frac{a}{n}||\vect{V}^{(i+1)}||_F^2 \label{lam2}
\end{equation}

  The proposal distribution $q(\vect{U},\vect{V} \mid \vect{U}^{(i)},\vect{V}^{(i)})$ used is the Auto-Regressive process with lag 1 such that
\begin{equation}
u_{ik}* = \alpha u_{ik} + z_{ik}^{(1)} \qquad v_{jk}* = \alpha u_{jk} + z_{jk}^{(2)}
\end{equation}
for $i = 1, 2, \ldots, n$,\  $j = 1, 2, \ldots, p$ and $k = 1, 2, \ldots, K$ where $u_{ik}$ and $v_{jk}$ are the elements of $\vect{U}$ and $\vect{V}$ respectively, $z_{ik}^{(1)}$ and $z_{jk}^{(2)}$ are iid normal random numbers independent of $\vect{U}$ and $\vect{V}$ with mean $0$ and variance $\sigma_1^2$ and $\sigma_2^2$ and $\alpha \in (-1, 1) \backslash \{ 0 \}$.

\section{Data Analysis}

  We implement the above algorithm in MovieLens small dataset which has been used with 751 users and 1616 movies. Initial choices of hyper parameters are set to $\lambda_1^{(0)} = \lambda_2^{(0)} = 10$.  From the Equation-\ref{lam1} and Equation-\ref{lam2}, we see that updates of $\lambda_1$ and $\lambda_2$ decrease with every iteration. Learning rate parameter $a$ is taken as $a = 5 \times 10^{-5}$.  The vectors $\vect{U}^{(0)}$ and $\vect{V}^{(0)}$ are initialized with iid normal random numbers such that the elements of $\vect{U}^{(0)}\vect{V}^{(0)T}$ have mean $3$ and variance $1$. 	We stop the iteration if changes in successive two hyper-parameters falls below $tol$. Here we take $tol = 10^{-5}$ and $\alpha = 0.9$ and $0.5$.   $(\sigma_1, \sigma_2)$ is taken as $(0.5, 0.5)$ and $(1, 1)$.  

\noindent{\textbf{Case 1:}}  When $\sigma_1 = \sigma_2 = 0.5$ and $\alpha = 0.9$, the hyper-parameters converge to $\hat{\lambda}_1 = 8.07549$ and $\hat{\lambda}_2 = 5.83292$.  

\noindent{\textbf{Case 2:}} When $\sigma_1 = \sigma_2 = 1$ and $\alpha$ is set to $0.5$, the hyper-parameters converge to $\hat{\lambda}_1 = 8.07058$ and $\hat{\lambda}_2 = 5.85553$.  

\begin{figure}[H]
 \begin{center}
  \subfigure[$\xi_{1}$]{\epsfig{file = 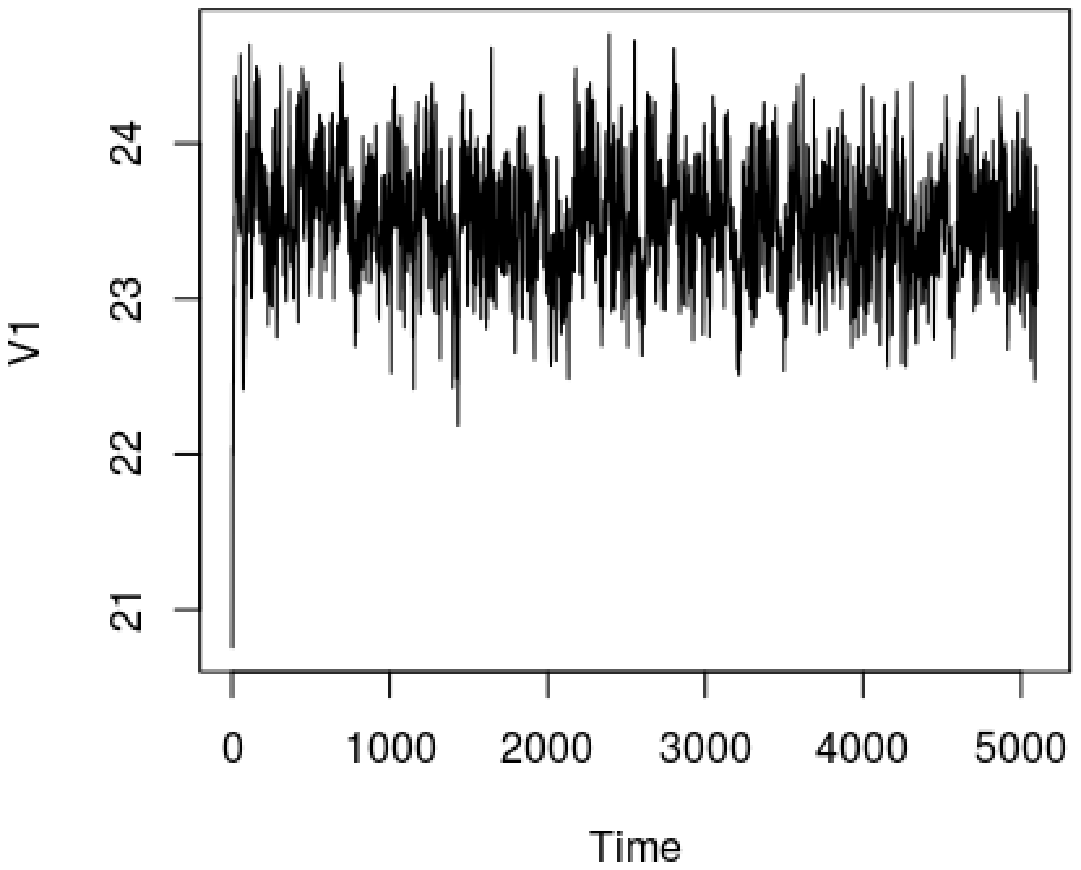, width = 7cm}}
  \subfigure[$\xi_{2}$]{\epsfig{file = 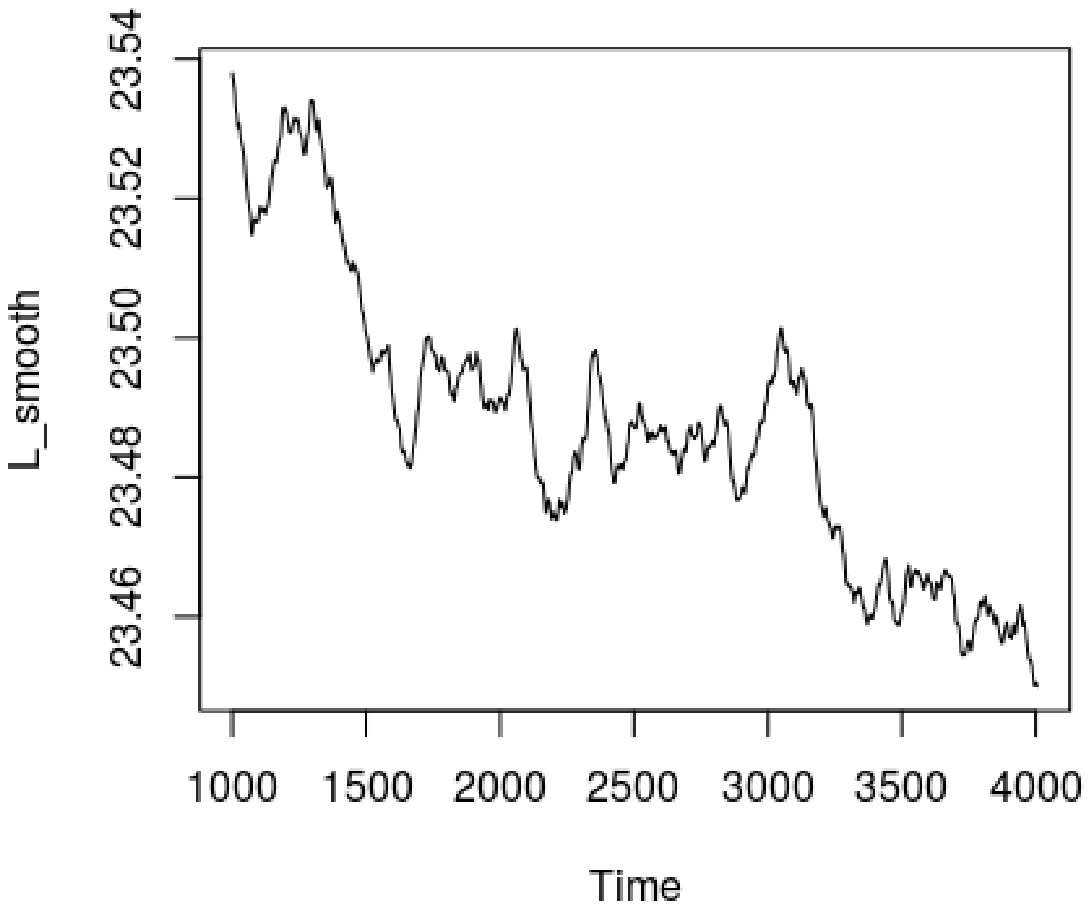, width = 7cm}}\\
\caption{Iterations vs Loss - Original and smoothened 1 \label{fig:2}}
\end{center}
\end{figure}

\begin{figure}[H]
   \subfigure[$\xi_{1}$]{\epsfig{file = 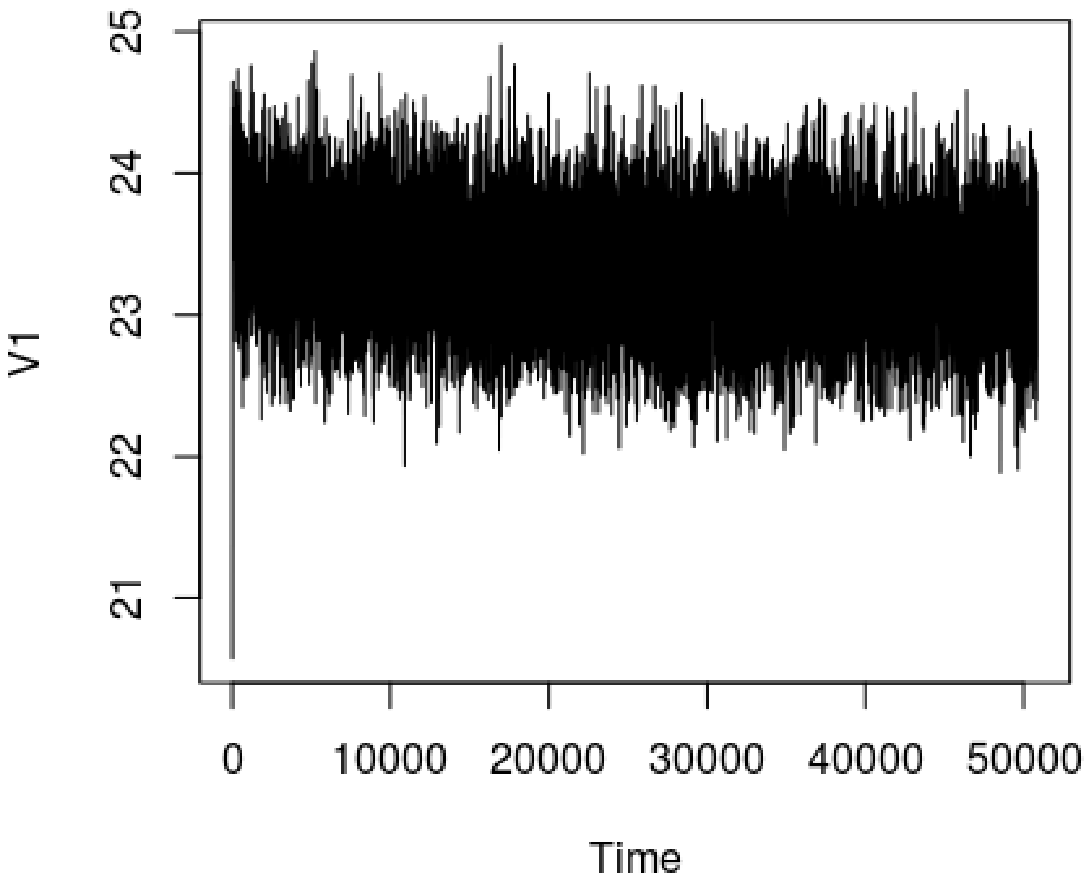, width = 7cm}}
   \subfigure[$\xi_{2}$]{\epsfig{file = 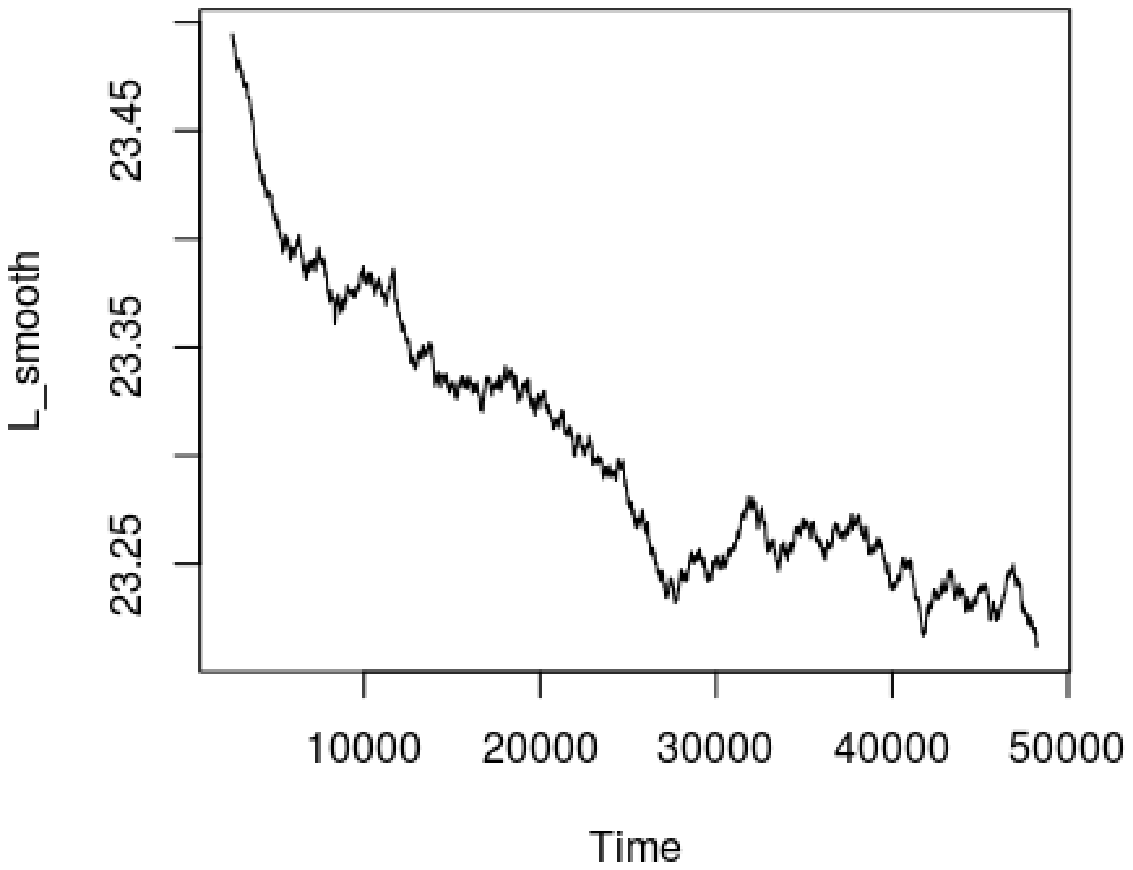, width = 7cm}}\\
 \caption{Iterations vs Loss - Original and smoothened 2 \label{fig:3}}	
\end{figure}

 Therefore we can see that the estimate is not much affected by initial choice of hyper-parameters.  In Figure-\ref{fig:1} and Figure-\ref{fig:2} we observe the behaviour of loss function over different iteration.  In both the figures, left hand side sub-figure is showing the original loss, whereas right hand side is smoothened loss with the moving-average. Value of $\alpha$ is taken as 0.9 in Figure-\ref{fig:2} whereas value of $\alpha$ is 0.5 for Figure-\ref{fig:3}. In both figures we observe the stability of loss for longer iteration.  We stop the algorithm if consecutive loss values are less than a centain tolerance.

  Using the estimated values of hyperparameters, the optimization algorithm was run and we obtain the validation RMSE of $1.13196$ and $1.13202$. However using the best hyper-parameter values found using grid search, the validation RMSE goes as below as $0.989$.  Therefore in terms of RMSE with the empirical bayes method is as good as the result found using stright grid search.  The advantage that we get in this approach is it provides much faster computation than the usual grid search method.  In fact estimated values of the hyperparameter can be taken as initial choice in grid search and can be made faster improvement of the result.  

\section{Conclusion}

  The proposed algorithm is fast and gives reasonable estimates of the hyper-parameters. The similar idea can be used with different other variations of CF to resolve the computer intensive hyper-parameter tuning.  This algorithm is also capable of giving the Bayes estimate of the parameters that are the factor matrices which can be used as an initial point for the optimization algorithm.  We need to deal with some hyper parameters which makes the algorithm complex to deal with. To ensure good convergence the choice of $a_n$ and the initial settings are crucial.


\nocite{chib1995understanding}

\bibliographystyle{chicago}
\bibliography{EBayes}


\end{document}